\documentclass[runningheads]{llncs}

\usepackage{eccv}

\usepackage{eccvabbrv}
\usepackage{multirow}
\usepackage{graphicx}
\usepackage{booktabs}

\usepackage[accsupp]{axessibility}  

\usepackage{hyperref}

\usepackage{orcidlink}

\begin{document}

\title{DreamMesh: Jointly Manipulating and Texturing Triangle Meshes for Text-to-3D Generation} 
\titlerunning{DreamMesh}

\author{Haibo Yang\inst{1,2 \ast} \and
Yang Chen\inst{3} \and
Yingwei Pan\inst{3} \and
Ting Yao\inst{3} \and \\
Zhineng Chen\inst{1,2 \dag} \and 
Zuxuan Wu\inst{1,2} \and
Yu-Gang Jiang\inst{1,2} \and
Tao Mei\inst{3}
}

\authorrunning{H. Yang et al.}


\institute{School of Computer Science, Fudan University \and Shanghai Collaborative Innovation Center of Intelligent Visual Computing \and
HiDream.ai Inc.
\\
\email{yanghaibo.fdu@gmail.com, \{c1enyang, pandy, tiyao\}@hidream.ai, \\
\{zhinchen, zxwu, ygj\}@fudan.edu.cn, tmei@hidream.ai}}

\maketitle

\let\thefootnote\relax\footnotetext{$^{\ast}$ This work was performed when Haibo Yang was visiting HiDream.ai as a research intern.}
\let\thefootnote\relax\footnotetext{$^{\dag}$ Corresponding author.}

\begin{abstract}
Learning radiance fields (NeRF) with powerful 2D diffusion models has garnered popularity for text-to-3D generation. Nevertheless, the implicit 3D representations of NeRF lack explicit modeling of meshes and textures over surfaces, and such surface-undefined way may suffer from the issues, e.g., noisy surfaces with ambiguous texture details or cross-view inconsistency. To alleviate this, we present DreamMesh, a novel text-to-3D architecture that pivots on well-defined surfaces (triangle meshes) to generate high-fidelity explicit 3D model. Technically, DreamMesh capitalizes on a distinctive coarse-to-fine scheme. In the coarse stage, the mesh is first deformed by text-guided Jacobians and then DreamMesh textures the mesh with an interlaced use of 2D diffusion models in a tuning free manner from multiple viewpoints. In the fine stage, DreamMesh jointly manipulates the mesh and refines the texture map, leading to high-quality triangle meshes with high-fidelity textured materials. 
Extensive experiments demonstrate that DreamMesh significantly outperforms state-of-the-art text-to-3D methods in faithfully generating 3D content with richer textual details and enhanced geometry. Our project page is available at \href{https://dreammesh.github.io}{https://dreammesh.github.io}.
  \keywords{Text-to-3D Generation \and Diffusion Models \and Triangle Meshes}
\end{abstract}

\section{Introduction}
\label{sec:intro}

Diffusion models~\cite{ho2020ddpm,ho2022classifier, zhu2024sd} have emerged as the basis of the powerful modern generative networks for producing realistic and diverse visual content (e.g., images and videos \cite{chen2019animating, chen2019mocycle, pan2017create}). In between, a massive leap forward has been attained in text-driven visual content generation tasks, e.g., text-to-image generation~\cite{imagen, Stable-diffusion, ramesh2022hierarchical, nichol2021glide, qian2024boosting} and text-to-video generation~\cite{ho2022video_diffusion, yang2022diffusion_video, zhang2024trip}. The success is attributed to several factors like billion-level multi-modal data and scalable denoising diffusion-based generative modeling. Nevertheless, it is not trivial to directly train a robust 3D-specific diffusion model for text-to-3D generation, since the paired text-3D data is relatively scarce, and 3D scenes have more complex geometric structures and multi-view visual appearances than 2D images.

\begin{figure}[t]
    \centering
    \includegraphics[width=1.0\textwidth]{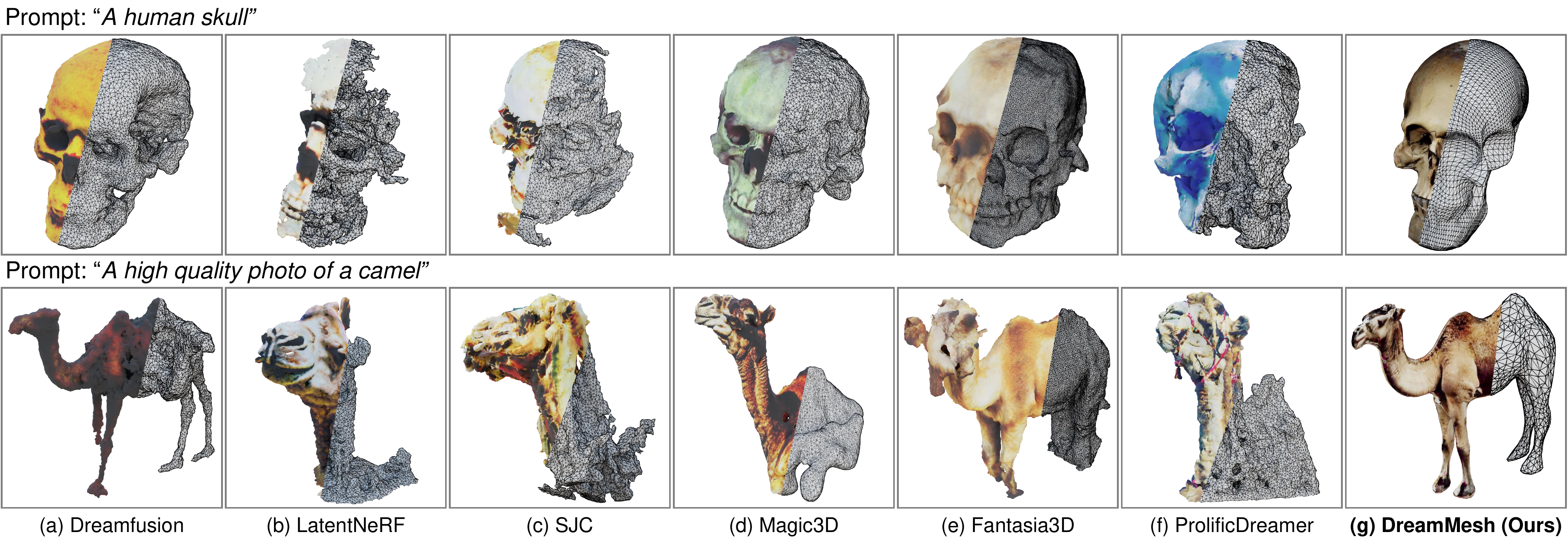}
    \caption{Existing methods~\cite{poole2022dreamfusion, metzer2022latent-nerf, wang2022SJC, lin2022magic3d, fantasia3d, wang2023prolificdreamer} mostly hinge on implicit or hybrid 3D representation and produce noisy surfaces. Instead, our DreamMesh pivots on completely~explicit 3D representation, yielding high-quality 3D meshes that exhibit clean, organized topology, devoid of any redundant vertices \&~faces.
    }
    \label{fig:teaser}
\end{figure}

The recent advance of Dreamfusion~\cite{poole2022dreamfusion} nicely sidesteps the requirement of massive paired text-3D data for text-to-3D generation task, and learns implicit 3D scene representation (NeRF~\cite{mildenhall2021nerf}) with only 2D diffusion models pre-trained over images. The core learning objective is to optimize implicit 3D scene with 2D observations of each sampled views derived from 2D diffusion models via Score Distillation Sampling (SDS). Despite having impressive quantitative results through SDS, qualitative analysis shows that such text-to-3D generation often results in cross-view inconsistency or ambiguous texture details due to the intrinsic bias of 2D diffusion priors. Later on, a series of efforts~\cite{lin2022magic3d, wang2022SJC, metzer2022latent-nerf, wang2023prolificdreamer, fantasia3d} have been dedicated to upgrading the 2D diffusion priors in SDS with 3D-aware knowledge, aiming to strengthen the capabilities to produce cross-view consistent 3D scene. 
Note that these text-to-3D works predominantly revolve around implicit 3D scene representation of density-based geometry with undefined surface boundaries. As shown in Figure~\ref{fig:teaser}, this surface-undefined framing easily leads to noisy extracted surfaces and over-saturated/over-smoothed textures. 
Moreover, the learnt 3D assets with implicit 3D scene fail to be directly integrated into graphics pipeline, and necessitate additional conversion from implicit to explicit 3D scene. The conversion might inject more noise over surfaces, thereby hindering the usage particularly in various high-quality 3D applications.

To address these challenges, our work shapes a new way to frame text-to-3D generation on the basis of completely explicit 3D scene representation of the ubiquitous and well-defined surface (triangle meshes). We propose a novel text-to-3D framework, namely DreamMesh, that executes the learning of textured triangle meshes into two stages. Specifically, in the first coarse stage, DreamMesh deforms the triangle meshes by text-guided Jacobians, obtaining globally smooth coarse mesh. Next, the corresponding coarse texture is attained through a tuning-free process with an interlaced use of pre-trained 2D diffusion models. In the second fine stage, DreamMesh jointly manipulates the coarse mesh and refines the coarse texture map. This scheme learns the surface and material/texture of explicit 3D representation in a coarse-to-fine fashion. Eventually the explicit 3D model by DreamMesh faithfully reflects the high-quality geometry (clean and organized topology) with rich texture details (see Figure \ref{fig:teaser}).

In summary, we have made the following contributions: (1) We novelly frame text-to-3D generation based on a completely explicit 3D scene representation of triangle meshes, which is shown capable of mitigating the issue associated with implicit 3D scenes and learning more smooth surfaces. (2) The exquisitely designed coarse-to-fine strategy pivoting on explicit 3D scene representation is shown able to facilitate the manipulation and texturing of triangle meshes. (3) The proposed DreamMesh has been analyzed and verified through extensive experiments over a comprehensive text-to-3D benchmark (T$^3$Bench~\cite{he2023t3bench}), demonstrating superior results when compared to state-of-the-art approaches.

\section{Related Works}
\textbf{Text-to-3D Generation.}
Recently, the text-to-3D generation has drawn increasing research attention. Pioneering works \cite{poole2022dreamfusion, wang2022SJC} utilize pre-trained 2D diffusion models to accomplish text-to-3D generation in a zero-shot fashion, mitigating the reliance on massive training data and becoming the mainstream. The key technique underpinning these methods is score distillation sampling (SDS), which enables distilling knowledge from the 2D diffusion model to optimize an underlying 3D representation (e.g., NeRF \cite{mildenhall2021nerf}) and showcases remarkable 3D generation capability. Subsequently, there has been a series of related works~\cite{metzer2022latent-nerf, chen2023control3d, lin2022magic3d, fantasia3d, yang20233dstyle, chen20233d, wang2023prolificdreamer, shi2023mvdream, zhu2023hifa, katzir2023noise, yu2023text, chen2024vp3d, tangdreamgaussian} that continue to refine and strengthen this methodology in different ways. For instance, Latent-NeRF \cite{metzer2022latent-nerf} and Control3D \cite{chen2023control3d} incorporate additional user-provided sketch mesh or image to guide the text-to-3D generation process. Magic3D \cite{lin2022magic3d} and Fantasia3D \cite{fantasia3d} upgrade the implicit NeRF representation used in DreamFusion to an implicit-explicit hybrid 3D representation (i.e., DMTet \cite{shen2021dmtet}) for SDS optimization on higher-resolution renderings. VP3D \cite{chen2024vp3d} leverages 2D visual prompt to boost text-to-3D generation. 

Although the aforementioned works can generate high-quality renderings, they all predominantly adopt implicit (NeRF) or implicit-explicit hybrid (DMTet) 3D representation. Integrating them into the mainstream graphics pipeline requires an additional conversion from implicit/hybrid 3D representation to widely used textured mesh. Unfortunately, this conversion may result in sub-optimal results due to the absence of explicit modeling meshes and textures during optimization, which prevents the usage of these methods in real-world deployments. In contrast, we formulate the text-to-3D generation process from a new perspective based on the completely explicit 3D representation of triangle meshes, leading to more clean $\&$ well-organized surfaces and photo-realistic textures that can be seamlessly compatible with existing 3D engines (e.g., Blender). 

\noindent \textbf{Text-Driven Shape Manipulation/Texturing.} 
Manipulating and texturing 3D shapes are key components in animation creation and computer-aided design pipelines, gaining a surge of interest in the literature. Classical approaches \cite{sorkine2004laplacian, sorkine2007rigid, huang2021arapreg, tang2022neural} perform shape manipulation by predicting mesh deformations from user-provided handles and cast this problem as an optimization task, where the source mesh is iteratively deformed to minimize the fitting error from the source to target shape. Instead of controlling the deformation through handle movements, a recent work \cite{TextDeformer} guides the deformation process solely from a text prompt by utilizing a pre-trained CLIP model \cite{radford2021CLIP} and differentiable rendering. Another direction of research focuses on text-driven 3D shape texturing that automatically generates textures for 3D bare meshes from the given text prompt. State-of-the-art methods \cite{TEXTure, chen2023text2tex} utilize pre-trained diffusion models (e.g., depth-to-image diffusion model and inpainting diffusion model) to ``paint'' the input bare mesh with generated textures. Unlike the aforementioned approaches, our work frames text-to-3D generation by jointly manipulating and texturing triangle meshes. Instead of formulating either a mesh deformation task or a mesh texturing problem in~\cite{TextDeformer, TEXTure, chen2023text2tex, yang20233dstyle, gao2024genesistex}, we uniquely look into the text-to-3D problem via generating explicit high-quality triangle meshes on the input text prompt.
\section{Approach}

\subsection{Preliminaries}
We first briefly review the typical score distillation sampling (SDS) method, and then discuss the relations and differences between our DreamMesh and related methods based on implicit-explicit hybrid 3D representation.
\label{Preliminaries}

\noindent \textbf{Score Distillation Sampling (SDS).} 
SDS is first introduced by Dreamfusion~\cite{poole2022dreamfusion} that leverages pre-trained text-to-image diffusion models to enable zero-shot text-to-3d generation. Specifically, DreamFusion employs Neural Radiance Fields (NeRF)~\cite{mildenhall2021nerf} to parameterize the implicit 3D scene as $\theta$. Next, a differentiable renderer is utilized to render an image $x$ from the 3D scene. In an effort to distill the knowledge of 2D diffusion model (e.g., Imagen~\cite{imagen}) into 3D scene, 
random noise $\epsilon$ is initially added to the image $x$:
\begin{equation}
    x_t=\sqrt{\Bar{\alpha_t}} x + \sqrt{1 - \Bar{\alpha_t}} \epsilon,
    \label{eq:noise_step}
\end{equation}
where $\epsilon \sim \mathcal{N}(0, I)$, and $\bar{\alpha}_t$ is a time-variant constant.
After that, DreamFusion employs the denoiser of diffusion model (parameterized as $\epsilon_{\phi}$) to estimate the added noise $\epsilon$ from the noisy image $x_t$. The 3D scene parameters $\theta$ are thus updated according to the per-pixel gradient of difference between the actual and predicted noise:
\begin{equation}\label{EqnSDS}
    \nabla_\theta \mathcal{L}_{\text{SDS}}(\phi,x) = 
    \mathbb{E}_{t,\epsilon} \left [ w(t)(\epsilon_{\phi}(x_t;y,t) - \epsilon)\frac{\partial x}{\partial \theta} \right ],
\end{equation}
where $w(t)$ is a weighting function and $y$ is the input text prompt. 
In this way, the pixel-level gradient is back-propagated to optimize the 3D scene, thereby driving the learnt 3D scene to resemble the input text prompt. 

\noindent \textbf{Implicit-Explicit Hybrid 3D Representations.}
Recent advances in text-to-3D generation predominantly employ implicit representations~\cite{mildenhall2021nerf} for modeling 3D scenes. A notable limitation of these techniques is the low-quality explicit mesh with noisy surfaces extracted from implicit fields. To alleviate this issue, several methods~\cite{fantasia3d,lin2022magic3d,wang2023prolificdreamer} employ implicit-explicit hybrid 3D representation (DMTet~\cite{shen2021dmtet}). Nevertheless, the meshes extracted from these DMTet-based methods still suffer from the problems such as excessive faces and poor topological structures. Such downside severely hinders the seamless integration of these meshes into traditional graphics rendering pipelines, thereby limiting their deployments in standard visualization and animation processes.
As an alternative, our DreamMesh pivots on completely explicit 3D representation (triangle meshes) for modeling 3D objects, thereby getting rid of the gap between implicit and explicit representations.
Formally, let $\mathcal{M}_0$ be a given base triangular mesh, which comprises a set of vertices $\mathcal{V} \in \mathbb{R}^{n\times 3}$ and triangular faces $T \in \mathbb{R}^{m\times 3}$. The base mesh can be a basic sphere, user-provided, or a low-quality mesh generated by 3D generative methods~\cite{li2023diffusionsdf, nichol2022pointe, jun2023shape-E, cheng2023sdfusion}. Through joint manipulation and texturing of triangle meshes, our DreamMesh learns and refines the 3D mesh and the corresponding mesh textures $\mathcal{T}$ that adhere to the given text prompt $y$.

\begin{figure}[t]
	\centering
	\includegraphics[width=1.0\textwidth]{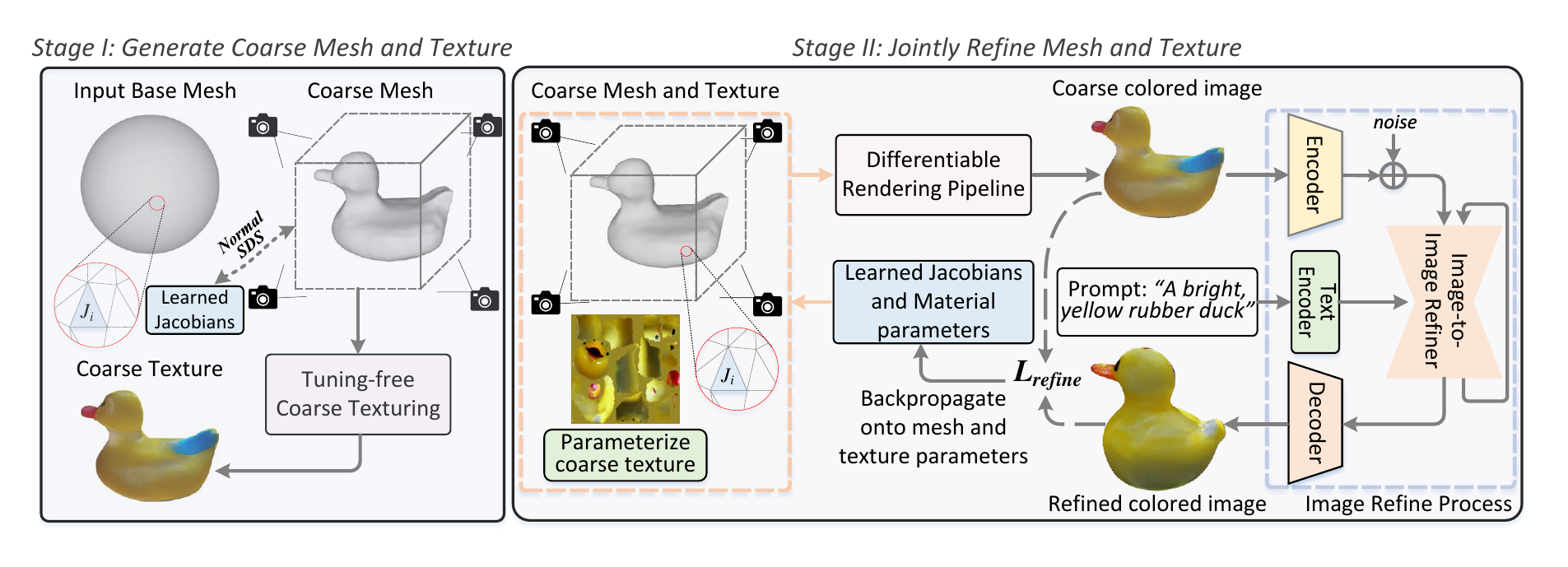}
	\caption{An overview of our DreamMesh that fully capitalizes on explicit 3D scene representation (triangle meshes) for text-to-3D generation in a coarse-to-fine scheme. In the first coarse stage, DreamMesh learns text-guided Jacobians matrices to deform a base mesh into the coarse mesh, and then textures it through a tuning-free process. In the second fine stage, both coarse mesh and texture are jointly optimized, yielding high-quality mesh with high-fidelity texture.
	}
	\label{fig:pipeline}
\end{figure}

\subsection{DreamMesh Optimization}

In this section, we elaborate our DreamMesh, which frames text-to-3D generation based on completely explicit 3D representation in a coarse-to-fine fashion. Figure~\ref{fig:pipeline} depicts the detailed framework, consisting of two stages: the coarse stage to produce coarse mesh and texture, and the fine stage to jointly refine mesh and texture with a diffusion-based 2D image-to-image refiner.

\noindent \textbf{Stage I: Generate Coarse Mesh and Texture.} In this stage, we aim to generate a coarse triangle mesh and textured map that respects the input text prompt. To achieve this, we first deform a base mesh into the coarse mesh and then texture it through a tuning-free process. Practically, we use Shap-E \cite{jun2023shape-E} outputs as the input base mesh. Note that Shap-E is trained on millions of paired text-3D data and can easily produce 3D objects with reasonable geometry. That makes Shpe-E an ideal choice to serve as the initial base mesh.

\textbf{Coarse Mesh Deformation.} Given a base triangular mesh $\mathcal{M}_0$, DreamMesh first learns to deform it into a target triangular mesh that faithfully matches the input text prompt. Technically, we formulate this learning process as the optimization of a displacement map $\mathbf{D}: \mathbb{R}^{3\times 3} \to \mathbb{R}^{3\times 3}$ over 
the vertices. Such piecewise linear mapping $\mathbf{D}$ of a mesh can be defined by assigning a new position to each one of the vertices: $\mathcal{V}_i \to D_i$. Nevertheless, direct optimization on the vertex positions of a triangular mesh can easily suffer from degeneracy and local minima, and thus overly distorts the original shape. To alleviate this issue, we take the inspiration from \cite{aigerman2022NJF, TextDeformer} and parameterize the mesh deformation by using a set of per-triangle Jacobians $J_i = \mathbf{D} \nabla^T_i(J_i\in\mathbb{R}^{3\times 3})$,
where $\nabla^T_i$ is the gradient operator of triangle $t_i \in T$. Given an arbitrary assignment of input matrix $M_i \in \mathbb{R}^{3\times3}$ for every triangle, we can achieve new vertex positions $\mathbf{D}^*$ whose Jacobians $J_i=\mathbf{D}^*\nabla^T_i$ are least-squares closest to $M_i$. And we can easily obtain the deformed vertex positions $\mathbf{D}^*$ by solving linear system:
\begin{equation}
\label{eq:poisson}
 \mathbf{D}^* = L^{-1}\mathcal{A}{\nabla^T} M,   
\end{equation}
where $\mathcal{A}$ is the mesh's mass matrix, $L$ is the mesh's Laplacian, 
and $M$ is the stacking of the input matrices $M_i$. Accordingly, the optimization of the deformation mapping ($\mathbf{D}$) can be interpreted as the optimization of the learnable Jacobians matrices ${M_i}$. In practice, we initialize these Jacobians matrices as identity matrices, where $\mathbf{D}^\ast$ is inherently established as the identity mapping. Please refer to~\cite{aigerman2022NJF} for the full technical details.

\textbf{Coarse Diffusion Guidance.}
To achieve text-driven deformation $\mathbf{D}^\ast$ that aligns with input text prompt, we exploit the powerful text-to-image diffusion model (Stable Diffusion \cite{Stable-diffusion}) as coarse diffusion guidance to facilitate Jacobians deformation.
Specifically, given the base mesh $\mathcal{M}_0$ and deformation mapping $\mathbf{D}^\ast$, we utilize a differentiable renderer $g_n$~\cite{Laine2020diffrast} to render a normal map $n$:
\begin{equation}
    n = g_n(\mathbf{D}^\ast(\mathcal{M}_0), c),
    \label{eq:image}
\end{equation}
where $c$ represents a camera pose that is arbitrarily sampled within the spherical coordinate system. 
Such random sampling strategy ensures a uniform distribution of camera poses across the sphere, providing comprehensive coverage and variability.
Next, during $t$-th timestep of diffusion process, we encode the rendered normal map $n$ into the latent space to obtain the latent code $z^{n}$,  and add Gaussian noise $\epsilon$ to get $z_t^{n}$.
The typical latent space SDS loss is thus utilized to optimize the deformation $\mathbf{D}^\ast$ by measuring the gradient w.r.t. $\mathbf{D}^\ast$ as:
\begin{equation}\label{sds_geometry}
    \nabla_{\mathbf{D}^\ast} \mathcal{L}_{\text{SDS}}(\phi, n) = \mathbb{E} \left [ w(t)(\hat{\epsilon}_{\phi}(z_t^{n};y,t) - \epsilon)\frac{\partial n}{\partial \mathbf{D}^\ast} \frac{{\partial z^{n}}}{\partial n}\right ],
\end{equation}
where $\hat{\epsilon}_{\phi}$ denotes denoiser in Stable Diffusion.
It is worthy to note that instead of using Stable Diffusion's image encoder, here we exploit a downsampled version of the normal map $n$ as the latent code~\cite{metzer2022latent-nerf, fantasia3d}, which leads to a faster convergence of $\mathbf{D}^\ast$. By randomly sampling views and backpropagating the gradient in Eq. (\ref{sds_geometry}) to the learnable parameters in $\mathbf{D}^\ast$, this way eventually achieves a target coarse mesh $\mathcal{M}_1 = \mathbf{D}^\ast(\mathcal{M}_0)$ that resembles the input text prompt.

\textbf{Coarse Texture Generation.} 
Next, we target for producing realistic coarse textures for the learnt coarse mesh $\mathcal{M}_1$. We apply a tuning-free approach to progressively generate coarse textures on the 3D triangle mesh with an interlaced use of pre-trained 2D diffusion models~\cite{TEXTure}. In particular, the texture is represented as an atlas ($\mathcal{T}_0$) learnt through UV mapping process~\cite{xatlas}.  At the initialization step, we use a differentiable renderer $\mathcal{R}$\cite{KaolinLibrary} to render a depth map $\mathcal{D}_0$ from an arbitrary initial viewpoint $v_0$, and use a pretrained depth-to-image diffusion model $\mathcal{M}_{depth}$~\cite{Stable-diffusion} conditioned on the rendered depth map to generate an initial colored image $I_0$. The generated image $I_0$ is then projected back to the texture atlas $\mathcal{T}_0$ to color the shape's visible parts from $v_0$. Following this initialization step, we iteratively change the viewpoint and alternatively use a pretrained inpainting diffusion model $\mathcal{M}_{paint}$~\cite{Stable-diffusion} or $\mathcal{M}_{depth}$ to generation new colored image. These colored images are projected back onto the initial texture $\mathcal{T}_0$. We repeat this process until a complete coarse texture map $\mathcal{T}_1$ is formed.

\noindent \textbf{Stage II: Jointly Refine Mesh and Texture.}\label{stage2} Recall that at the first coarse stage, the optimization process of coarse mesh deformation solely focuses on the primary mesh irrespective of any texture. Such process might inevitably simulate textured results and lead to excessive modifications of meshes. Meanwhile, the coarse texture generation in first stage also encounters the inconsistency issue across all viewpoints. We speculate that the sub-optimal texturing results might be caused by the tuning-free texturing strategy that performs progressive texture mapping starting from a singular viewpoint, and it is non-trivial to maintain both local and global consistency. To alleviate these, we novelly devise a fine stage to jointly optimize both the mesh and texture with fine diffusion guidance derived from a pretrained diffusion-based image refiner \cite{podell2023sdxl}.
This innovative stage establishes a symbiotic relationship between the learnt meshes and textures, harmonizing the enhancement that contributes to the overall realism and consistency of synthetic 3D content.

Technically, in the fine stage, we adopt the same mesh deformation methodology as in the coarse stage, i.e., the optimization of Jacobian Matrices, to refine the coarse mesh $\mathcal{M}_1$.
However, different from the tuning-free texturing strategy in coarse stage, here we concurrently parameterize the coarse texture map $\mathcal{T}_1$, and trigger a joint optimization of $\mathcal{M}_1$ and $\mathcal{T}_1$ in the fine stage. 
By doing so, we employ a differentiable rendering pipeline $g$, which includes a sequence of mesh operations, a rasterizer, and a deferred shading stage~\cite{Hasselgren2021} to render a coarse colored image $x_{coarse}$ derived from the deformering mesh $\mathcal{M}_1$ and parameterized texture map $\mathcal{T}_1$, conditioned on a random camera pose $c$:
\begin{equation}
    x_{coarse} = g(\mathbf{D}^\ast_{fine}(\mathcal{M}_1), \mathcal{T}_1, c).
    \label{eq:refine}
\end{equation}

\textbf{Fine Diffusion Guidance.}
Our fine stage necessitates intricate adjustments to the coarse mesh structure and dedicated efforts to enhance texture consistency.
One natural way to optimize such fine process is to use the same coarse diffusion guidance (SDS loss) as in coarse stage for supervision. Nevertheless, such way will result in various artifacts like oversaturated color blocks. Instead, we excavate the fine diffusion guidance by additionally refining rendered coarse colored image $x_{coarse}$ with diffusion-based image refiner $\mathcal{E}$~\cite{podell2023sdxl}. This refined colored image $x_{refine} = \mathcal{E}(x_{coarse}, y)$ is further utilized to guide the joint optimization of mesh and texture through Mean Squared Error (MSE) loss:
\begin{equation}
\begin{aligned}
	&\mathcal{L}_{refine}(\mathbf{D}^\ast_{fine}, \mathcal{T}_1) = {||x_{refine} - x_{coarse}||}^2_2 \\
	&= {||\mathcal{E}(g(\mathbf{D}^\ast_{fine}(\mathcal{M}_1), \mathcal{T}_1, c), y) - g(\mathbf{D}^\ast_{fine}(\mathcal{M}_1), \mathcal{T}_1, c)||}^2_2.
\end{aligned}
\end{equation}
By minimizing this objective, our DreamMesh enforces the rendered image $x_{coarse}$ visually similar as the refined image $x_{refine}$ that faithfully matches with text prompt, thereby yielding high-quality mesh with high-fidelity texture map.

\textbf{Discussion.}
Some related works \cite{TextDeformer, TEXTure, chen2023text2tex} also explore text-driven mesh deformation or texturing, while our DreamMesh targets for a different task of text-to-3D generation. Instead of a simple combination of existing mesh deformation $\&$ texturing techniques (see degenerated results in Fig. \ref{fig:naive}), our DreamMesh novelly upgrades mesh deformation with geometry-aware supervision and further bridges both worlds by jointly optimizing mesh and texture with a fine-grained image-to-image refiner in fine stage.

\section{Experiments}
\subsection{Experimental Settings}
\textbf{Implemention Details.}
At the coarse stage, we utilize an Adam optimizer with a learning rate of $2\times10^{-3}$ and render $12$ normal maps per iteration. The coarse textures are generated from $10$ different viewpoints. In the fine stage, we set the learning rate as $2\times10^{-3}$ for mesh optimization and $1\times10^{-2}$ for texture material refinement. The diffusion-based image refiner performs denoising operations over 15 steps to produce refined images. The whole experiment is conducted on a single NVIDIA RTX 3090 GPU, and the learning process of each sample takes approximately $30$ minutes.

\noindent \textbf{Dataset.}
Most existing text-to-3D generation methods solely perform case studies and user surveys for evaluation, but lack quantitative assessment due to the absence of standard benchmark. Thanks to the newly released T$^3$Bench~\cite{he2023t3bench}, we perform quantitative comparisons over this first comprehensive benchmark for text-to-3D generation. Specifically, T$^3$Bench contains 100 test prompts for each of three categories (single object, single object with surroundings, and multiple objects). Two automatic metrics are designed to evaluate the subjective quality and textual alignment, based on the rendered multi-view images generated from 3D models. The quality metric combines multi-view text-image scoring with regional convolution to assess both quality and view consistency. The alignment metric first employs 3D-to-text caption model to achieve multi-view captions and then leverages Large Language Model (GPT-4) to merge captions into 3D caption for text-3D alignment assessment.

\noindent \textbf{Compared Methods.}
To empirically verify the merit of our DreamMesh, we include six state-of-the-art approaches for comparison. Specifically, \textbf{DreamFusion}~\cite{poole2022dreamfusion}, \textbf{LatentNeRF}~\cite{metzer2022latent-nerf}, and \textbf{SJC}~\cite{wang2022SJC} fully hinge on implicit 3D representation (NeRF~\cite{mildenhall2021nerf}) for text-to-3D generation. \textbf{Magic3D}~\cite{lin2022magic3d} upgrades DreamFusion with additional stage that capitalizes on implicit-explicit hybrid 3D representation (DMTet~\cite{shen2021dmtet}) to enhance texture details. \textbf{Fantasia3D}~\cite{fantasia3d} disentangles geometry and appearance modeling in two stages, i.e., first generating meshes based on DMTet and then leveraging Bidirectional Reflectance Distribution Function (BRDF) to produce textures.
\textbf{ProlificDreamer}~\cite{wang2023prolificdreamer} extends Magic3D by generalizing SDS in the variational formulation, aiming to alleviate the restricted diversity issue rooted in typical SDS.

\subsection{Quantitative Results}

\label{5.3}
Table~\ref{tb:t3bench} summarizes the quantitative performance comparisons over three categories of T$^3$Bench benchmark between our DreamMesh and six state-of-the-art approaches. Overall, for each category of text prompts, our DreamMesh consistently achieves better performances against the existing methods across all metrics, including both implicit 3D representation-based methods (Dreamfusion, LatentNeRF, SJC) and implicit-explicit hybrid 3D representation-based methods (Magic3D, Fantasia3D, ProlificDreamer). In particular, the average score of quality and alignment of DreamMesh can reach 54.7\%, 48.7\%, and 39.2\% for each category, which leads to the absolute improvement of 5.3\%, 3.9\%, and 3.4\% against the best competitor ProlificDreamer. The results clearly demonstrate the key advantage of joint manipulating and texturing based on completely explicit 3D representation to facilitate text-to-3D generation.

\begin{table}[t]
	\centering  
	\caption{Quantitative comparison between our DreamMesh and various text-to-3D generation approaches on T$^3$Bench~\cite{he2023t3bench} benchmark.}
	\resizebox{\linewidth}{!}{
		\begin{tabular}{l|c|ccc|ccc|ccc}
			\toprule[2pt]
			& \multirow{2}{*}{3D Representation}  & \multicolumn{3}{c}{\emph{Single Object}} & \multicolumn{3}{c}{\emph{Single Object with Surroundings}} & \multicolumn{3}{c}{\emph{Multiple Objects}}  \\
			& & Quality & Alignment & Average  & Quality & Alignment & Average  & Quality & Alignment & Average   \\ \midrule
			Dreamfusion~\cite{poole2022dreamfusion} & \multirow{3}{*}{\shortstack{Implict\\Representation\\ (NeRF~\cite{mildenhall2021nerf})}} & 24.9 & 24.0 & 24.4 & 19.3 & 29.8 & 24.6 & 17.3 & 14.8 & 16.1  \\
			LatentNeRF~\cite{metzer2022latent-nerf} & & 34.2 & 32.0 & 33.1 & 23.7 & 37.5 & 30.6 & 21.7 & 19.5 & 20.6  \\
			SJC~\cite{wang2022SJC} & & 26.3 & 23.0 & 24.7 & 17.3 & 22.3 & 19.8 & 17.7 & 5.8  & 11.7  \\
			\midrule
			Magic3D~\cite{lin2022magic3d} & \multirow{3}{*}{\shortstack{Implicit-Explicit\\Hybrid Representation\\ (DMTet~\cite{shen2021dmtet})}}& 38.7 & 35.3 & 37.0 & 29.8 & 41.0 & 35.4 & 26.6 & 24.8 & 25.7  \\
			Fantasia3D~\cite{fantasia3d} & & 29.2 & 23.5 & 26.4 & 21.9 & 32.0 & 27.0 & 22.7 & 14.3 & 18.5  \\
			ProlificDreamer~\cite{wang2023prolificdreamer} &  & 51.1 & 47.8 & 49.4 & 42.5 & 47.0 & 44.8 & 45.7 & 25.8 & 35.8  \\
			\midrule
			\multirow{2}{*}{\textbf{DreamMesh (Ours)}} & \multirow{2}{*}{\shortstack{Explicit Representation\\(Triangle Mesh)}} & \multirow{2}{*}{\textbf{55.6}} & \multirow{2}{*}{\textbf{53.8}} &  \multirow{2}{*}{\textbf{54.7}} &  \multirow{2}{*}{\textbf{43.1}} & \multirow{2}{*}{\textbf{54.3}} &  \multirow{2}{*}{\textbf{48.7}} & \multirow{2}{*}{\textbf{47.6}} & \multirow{2}{*}{\textbf{30.8}} & \multirow{2}{*}{\textbf{39.2}} \\
			& & &  & & & & & & & \\
			\bottomrule[2pt]
		\end{tabular}
	}
	\label{tb:t3bench}
\end{table}

More specifically, Dreamfusion enables a zero-shot solution of optimizing implicit 3D representation with 2D diffusion priors, yielding promising results even under challenging prompts in the categories of ``single object with surroundings'' and ``multiple objects''). SJC remoulds Dreamfusion by performing score jacobian chaining within the voxel version of NeRF~\cite{DVGO, Chen2022ECCV_TensoRF}, which easily results in degraded 3D models with a significant amount of floating density. In contrast, LatentNeRF upgrades Dreamfusion with a coarse-to-fine paradigm by using the implicit 3D representations of latent NeRF that is more tailored to the 2D latent diffusion model (Stable Diffusion~\cite{Stable-diffusion}), thereby leading to clear performance boosts.
Furthermore, compared to aforementioned three methods that solely exploit implicit 3D representations, Magic3D exhibits better performances by capitalizing on implicit-explicit hybrid 3D representation (DMTet) to learn textured 3D mesh. Fantasia3D also explores DMTet in a decoupled mesh generation stage and leverages BRDF for texture generation that emphasizes rich object textures, while it fails to create complex and high-fidelity meshes (e.g., ``multiple objects'' category).
ProlificDreamer further boosts up the performances by upgrading SDS of Magic3D in variational formulation to address the restricted diversity issue. Nevertheless, ProlificDreamer still relies on implicit-explicit hybrid 3D representation (DMTet) and commonly suffers from noisy surfaces with over-smoothed textures. In contrast, our DreamMesh completely eliminates the use of implicit 3D representation and achieves the best performances through coarse-to-fine strategy pivoting on explicit 3D representation of triangle meshes.

\subsection{Qualitative Results}

\begin{figure}[t]
	\centering
	\includegraphics[width=1.0\textwidth]{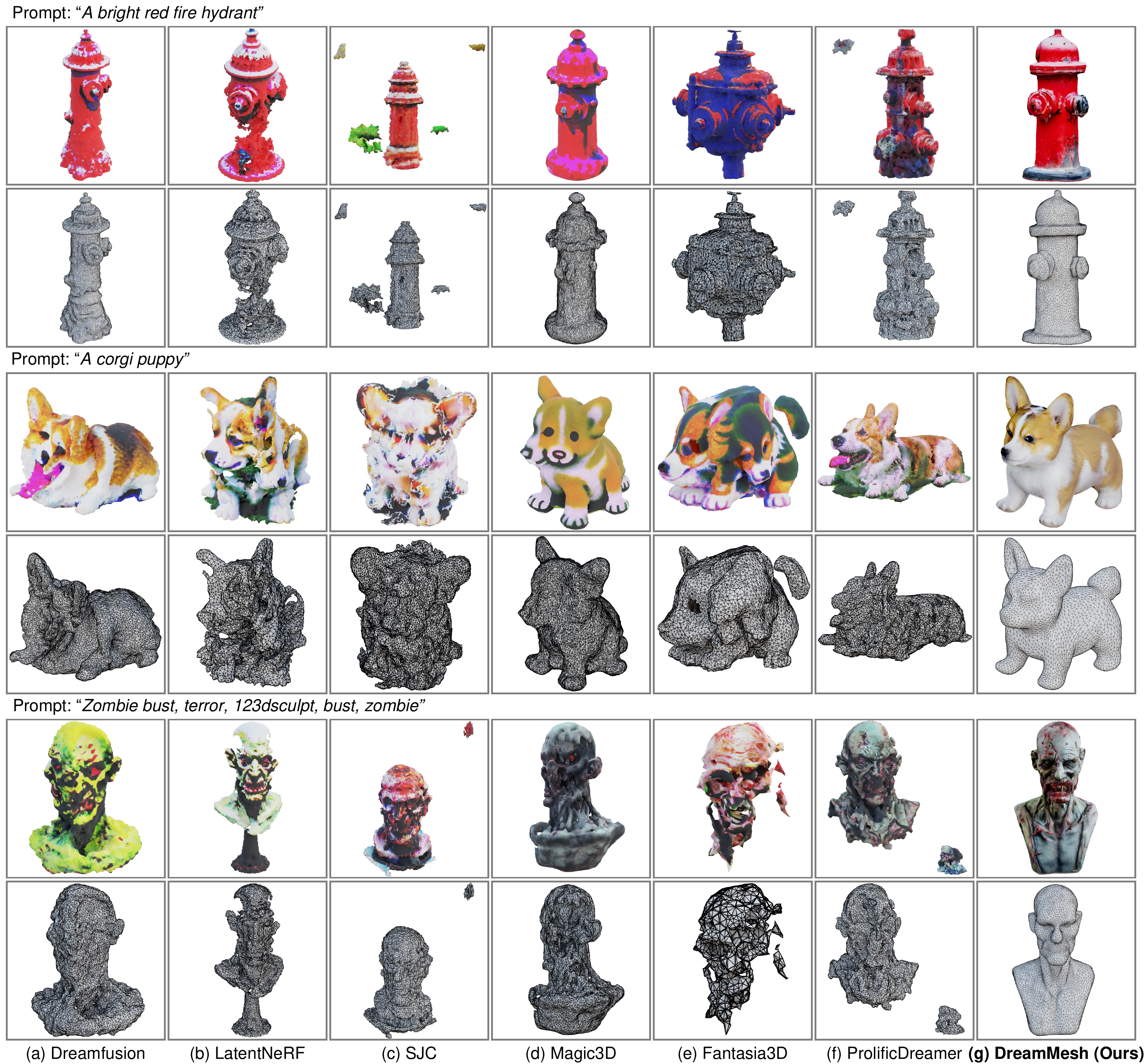}
	\caption[]{
		Qualitative comparison of texture and wireframe results (rendering in Blender) between our DreamMesh and other baseline methods.
		\label{fig:results}}
\end{figure}

As indicated by these exemplar results in Figure~\ref{fig:results}, all the methods can generate somewhat reasonable meshes and textures, while our DreamMesh can synthesize higher quality meshes with richer textures that faithfully adhere to text prompts by pivoting on completely explicit 3D representation. For instance, given the first text prompt \textit{``A bright red fire hydrant''}, the implicit 3D representation-based methods (Dreamfusion, LatentNeRF, SJC) produce noisy surfaces and simple textures with obvious deformation. By exploring implicit-explicit hybrid 3D representation for text-to-3D generation, Magic3D, Fantasia3D, and ProlificDreamer further yield more complete and accurate meshes. Nevertheless, these meshes generated via DMTet are inherently complex, necessitating excessive faces and vertices, resulting in unsatisfactory triangle topologies. In contrast, our DreamMesh novelly manipulates and textures well-defined surface (triangle meshes), leading to high-quality textured meshes that reflect clean and well-organized topology with neatly arranged vertices, edges, and faces.

\subsection{Experimental Analysis}
\label{sec: experimental_analysis}

\begin{figure}[t]
	\begin{center}
		\includegraphics[width=0.8\textwidth]{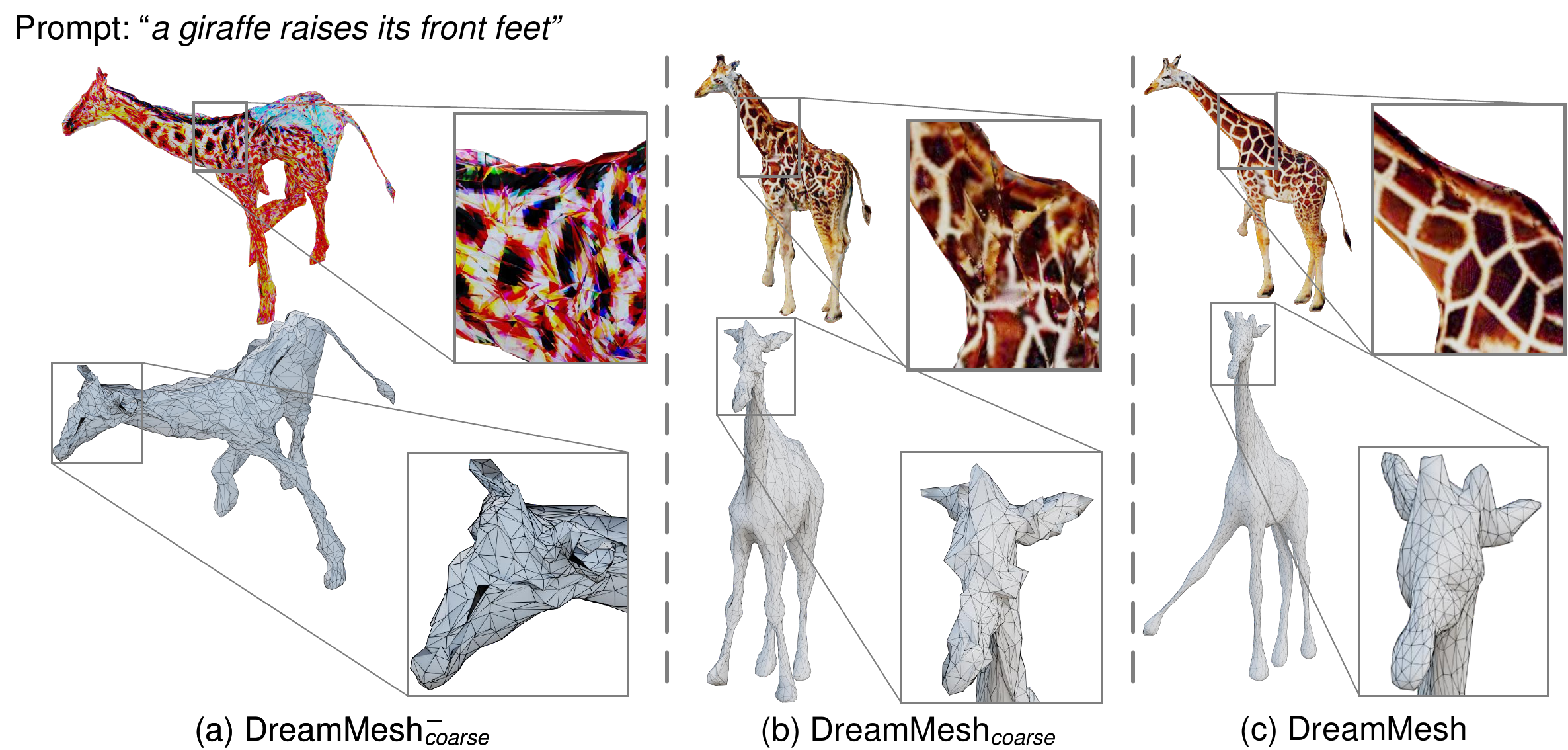}
	\end{center}
	\caption{Ablation study of our DreamMesh given the same text prompt. DreamMesh$_{coarse}^{-}$ is a degraded version of coarse stage that jointly optimizes mesh deformation and textures via SDS. DreamMesh$_{coarse}$ is the complete coarse stage that decouples the learning of coarse meshes and textures. DreamMesh is our full run with both coarse and fine stages.}
	\label{fig:ablation}
\end{figure}

\textbf{Ablation Studies.}
In an effort to study the effectiveness of each design in our DreamMesh, we depict the qualitative results of several ablated runs in Figure~\ref{fig:ablation}. 
DreamMesh$_{coarse}^{-}$ is an alternative version of our coarse stage by simultaneously optimizing mesh and texture from scratch via typical SDS, which easily leads to sub-optimal results. By decoupling mesh deformation and texturing into two separate processes, DreamMesh$_{coarse}$ further enhances the quality of textured meshes. The results validate the effectiveness of decoupled mesh and texture modeling in coarse stage. Nevertheless, DreamMesh$_{coarse}$ still suffers from some overly distorted meshes. DreamMesh upgrades DreamMesh$_{coarse}$ with an additional fine stage to jointly manipulate coarse meshes and refine coarse textures, yielding higher-quality fine meshes and textures.

\begin{figure}[t]
	\centering
	\includegraphics[width=0.8\textwidth]{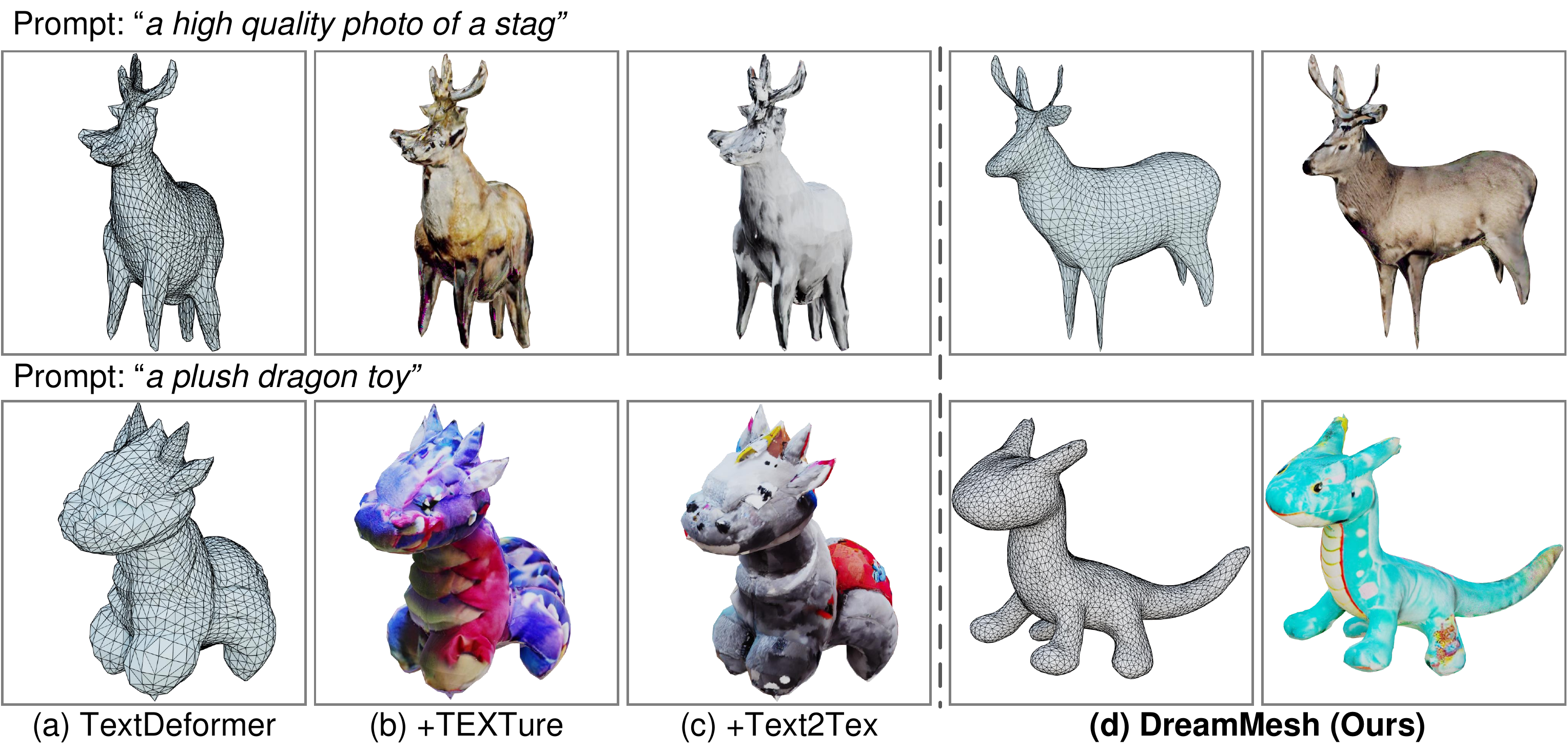}
	\caption{Comparisons between our DreamMesh and the integration of the state-of-the-art text driven mesh deformation technique \cite{TextDeformer} and the advanced texturing methods of TEXTure \cite{TEXTure} or Text2Tex \cite{chen2023text2tex} for text-to-3D generation.}
	\label{fig:naive}
\end{figure}

To further verify the leverage of explicit 3D representation for text-to-3D generation, we also compare our DreamMesh with the run of integrating the state-of-the-art text driven mesh deformation technique \cite{TextDeformer} and the advanced texturing methods of TEXTure \cite{TEXTure} or Text2Tex \cite{chen2023text2tex}. Figure \ref{fig:naive} shows the comparisons. It is not surprising that TextDeformer and TextDeformer+TEXTure/Text2Tex produce unsatisfactory surfaces and textures since these methods are not particularly tailored for text-to-3D generation. Our DreamMesh, in comparison, introduces more powerful diffusion model for mesh deformation and benefits from the coarse-to-fine optimization scheme, making the mesh more clean and the texture more realistic. Moreover, we conduct quantitative evaluations for the aforementioned runs on a random subset of T$^3$bench (50 prompts) and Table~\ref{tab:tab_abl} lists the results. The runs of TextDeformer+TEXTure/Text2Tex indicate poorest quality and alignment, again revealing the weakness of simple combination of mesh deformation and texture techniques for text-to-3D generation. DreamMesh$_{coarse}^{-}$ is inferior to DreamMesh$_{coarse}$, showing that entangled optimizing mesh and texture from scratch is more challenging. Finally, DreamMesh achieves the best performance, validating the effectiveness of our exquisitely designed coarse-to-fine strategy.

\begin{table}[t]
	\centering  
	\caption{Quantitative comparisons on the T$^3$Bench subset.}
	\resizebox{1.0\linewidth}{!}{
		\begin{tabular}{l|c|c|c|c|c}\toprule[1.0pt]
			& ${\rm TextDeformer} + {\rm TEXTure}$ & ${\rm TextDeformer} + {\rm Text2Tex}$ & ${\rm DreamMesh}_{coarse}^{-}$ & ${\rm DreamMesh}_{coarse}$ &  ${\rm DreamMesh}$ \\ 
			\midrule
			Quality & 20.2 & 19.3 & 24.4   &  40.5  & \textbf{47.1}     \\
			Alignment & 19.0 & 16.5 & 22.0   &  38.5  & \textbf{45.5}     \\
			Average & 19.6 & 17.9 & 23.2   &  39.5  & \textbf{46.3}     \\
			\bottomrule[1.0pt]
		\end{tabular}
	}
	\label{tab:tab_abl}
\end{table}

\begin{figure}[t]
    \begin{center}
    \includegraphics[width=1.0\textwidth]{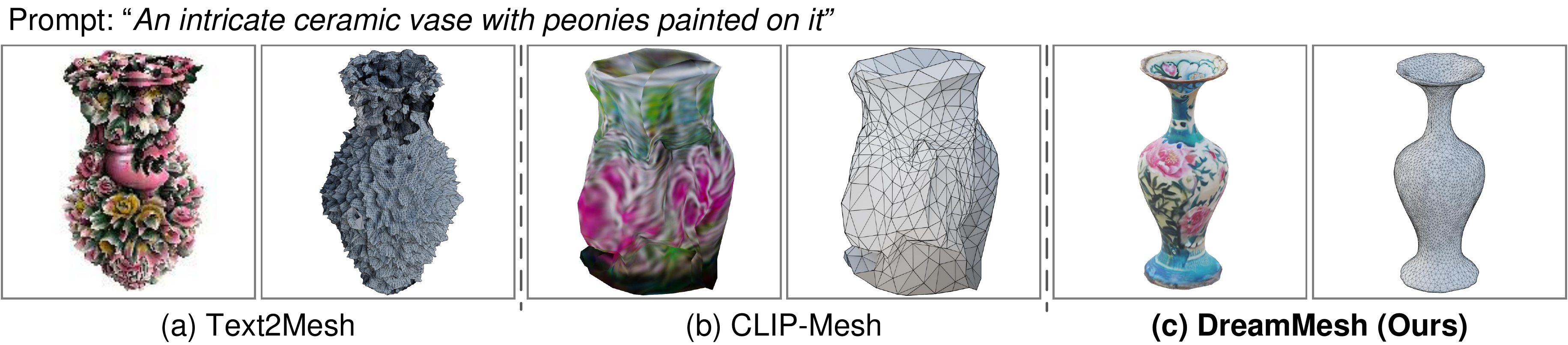}
    \end{center}

    \caption{Qualitative comparison of texture and wireframe results (rendering in Blender) against Text2Mesh and CLIP-Mesh.}

    \label{fig:mesh-based method}
\end{figure}

\begin{figure}[t]
    \begin{center}
    \includegraphics[width=1.0\textwidth]{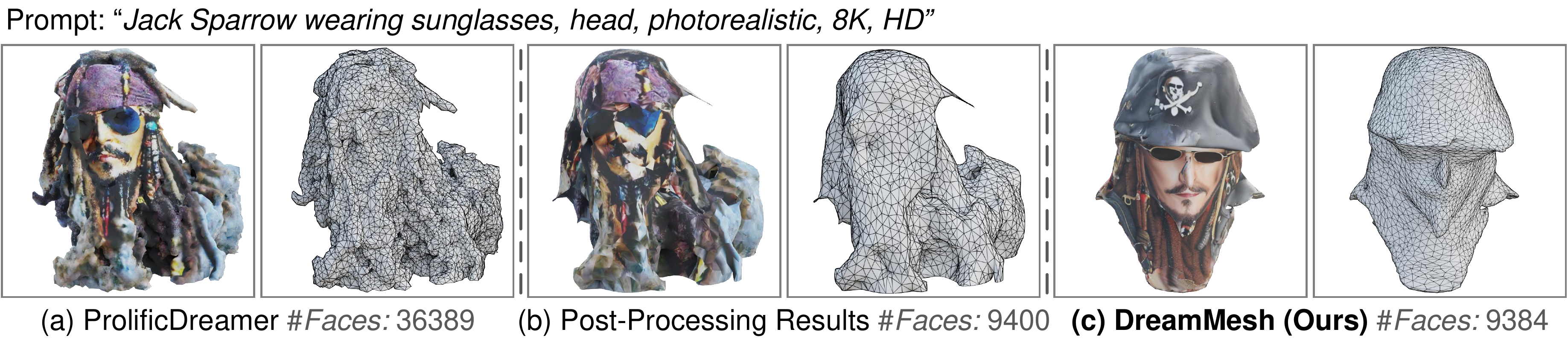}
    \end{center}

    \caption{Qualitative comparison of texture and wireframe results (rendering in Blender) against ProlificDreamer with additional manual post-processing.}

    \label{fig:post-process mesh}
\end{figure}

\begin{figure}[t]
   \begin{center}
    \includegraphics[width=1.0\textwidth]{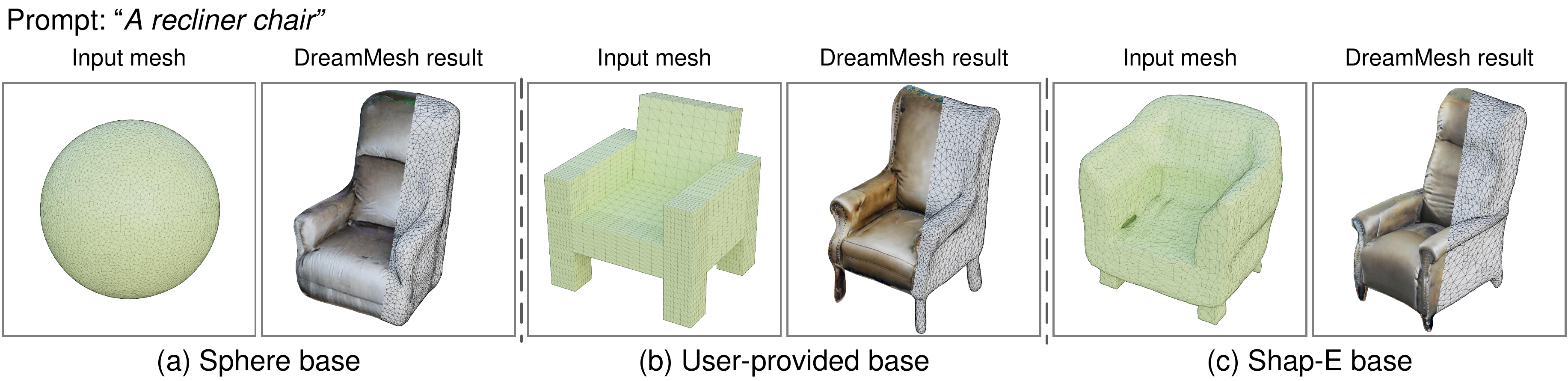}
    \end{center}

    \caption{Text-to-3D generation results (rendering in Blender) of DreamMesh with different input base meshes.}
    
    \label{fig:different input mesh}
\end{figure}

\noindent \textbf{Comparison against Typical Mesh-based Methods without Diffusion Model.}
It is worthy to note that some early attempts (e.g., Text2Mesh \cite{Michel_2022_CVPR_text2mesh} and CLIP-Mesh \cite{khalid2022clipmesh}) also explore explicit 3D representation for text-to-3D generation, while no powerful diffusion model is adopted. Figure~\ref{fig:mesh-based method} shows the qualitative comparison between our DreamMesh against Text2Mesh and CLIP-Mesh. In general, our DreamMesh significantly outperforms the conventional mesh-based methods with regard to both mesh/texture quality and text-3D alignment. This confirms the merit of exploiting explicit 3D representation for text-to-3D generation conditioned on powerful 2D diffusion priors.

\noindent \textbf{Comparison against ProlificDreamer with Manual Post-processing.}
Recall that both implicit and implicit-explicit hybrid representations of NeRF and DMTet can be transformed into explicit meshes by using the Marching Cubes and Marching Tetrahedral layer respectively. However, such automatic conversion commonly injects more noise over surfaces and leads to extremely complex meshes containing a large number of vertices, edges, and faces (e.g., 36,389 faces in Figure~\ref{fig:post-process mesh} (a)). 
To alleviate this issue, manual post-processing (e.g., cleaning, smoothing and simplification) can be employed to improve the mesh quality, while losing many geometric details (e.g., the missing of nose in Figure~\ref{fig:post-process mesh} (b)). In contrast, as shown in Figure~\ref{fig:post-process mesh} (c), our DreamMesh manages to achieve high-quality textured meshes that exhibit clean and organized topology.

\noindent \textbf{Comparison with Different Base Meshes.} Our DreamMesh is able to perform text-to-3D generation based on different kinds of input base meshes. For example, given the input text prompt ``A recliner chair'', we can generate it directly from a sphere. Alternatively, users can quickly create a rough 3D shape that approximately aligns with text prompt in 3D engines (e.g., Blender). Such user-provided rough mash can be fed into DreamMesh. Additionally, we can take the low-quality mesh automatically generated by 3D generative models (e.g., Shap-E~\cite{jun2023shape-E}) as the inputs. As shown in Figure~\ref{fig:different input mesh}, all text-to-3D generation results with different base meshes (i.e., basic sphere, user-provided shape, or Shap-E outputs) can produce higher-quality 3D assets, which generally demonstrate the generalization ability of our DreamMesh.

\section{Conclusions}
In this paper, we propose DreamMesh, a novel framework for text-to-3D generation that fully relies on explicit 3D representations in a coarse-to-fine manner. Specifically, in the coarse stage, we leverage neural jacobian fields to deform a triangle mesh and then texture the generated coarse mesh through a tuning-free process with an interlaced use of pre-trained 2D diffusion models. In the fine stage, we jointly refine the coarse mesh and texture to produce high-quality 3D model with rich texture details and enhanced 3D geometry. We evaluate our proposal on T$^3$Bench benchmark and demonstrate its superiority over state-of-the-art techniques through both qualitative and quantitative comparisons.

\textbf{Limitations and Broader Impact.} DreamMesh may suffer from multi-face Janus problems in some cases due to the limited 3D awareness of the prior 2D diffusion model. Finetuning the diffusion model on 3D data might alleviate this problem. Since the generated meshes can be seamlessly compatible with existing 3D engines, DreamMesh has great potential to displace creative workers via automation, which may enable growth for the creative industry. Nevertheless, it could also be potentially applied to unexpected scenarios such as generating fake and malicious content, which needs more caution. 

\noindent\textbf{Acknowledgement:} \quad This work is supported by the National Natural Science Foundation of China (No. 32341012 and No. 62172103).

%
%
\bibliographystyle{splncs04}
\bibliography{egbib}

\appendix
\section*{Appendix}
\section{Application of DreamMesh in 3D Rendering Pipeline}
\begin{figure}[t]
	\centering
	\includegraphics[width=1.0\textwidth]{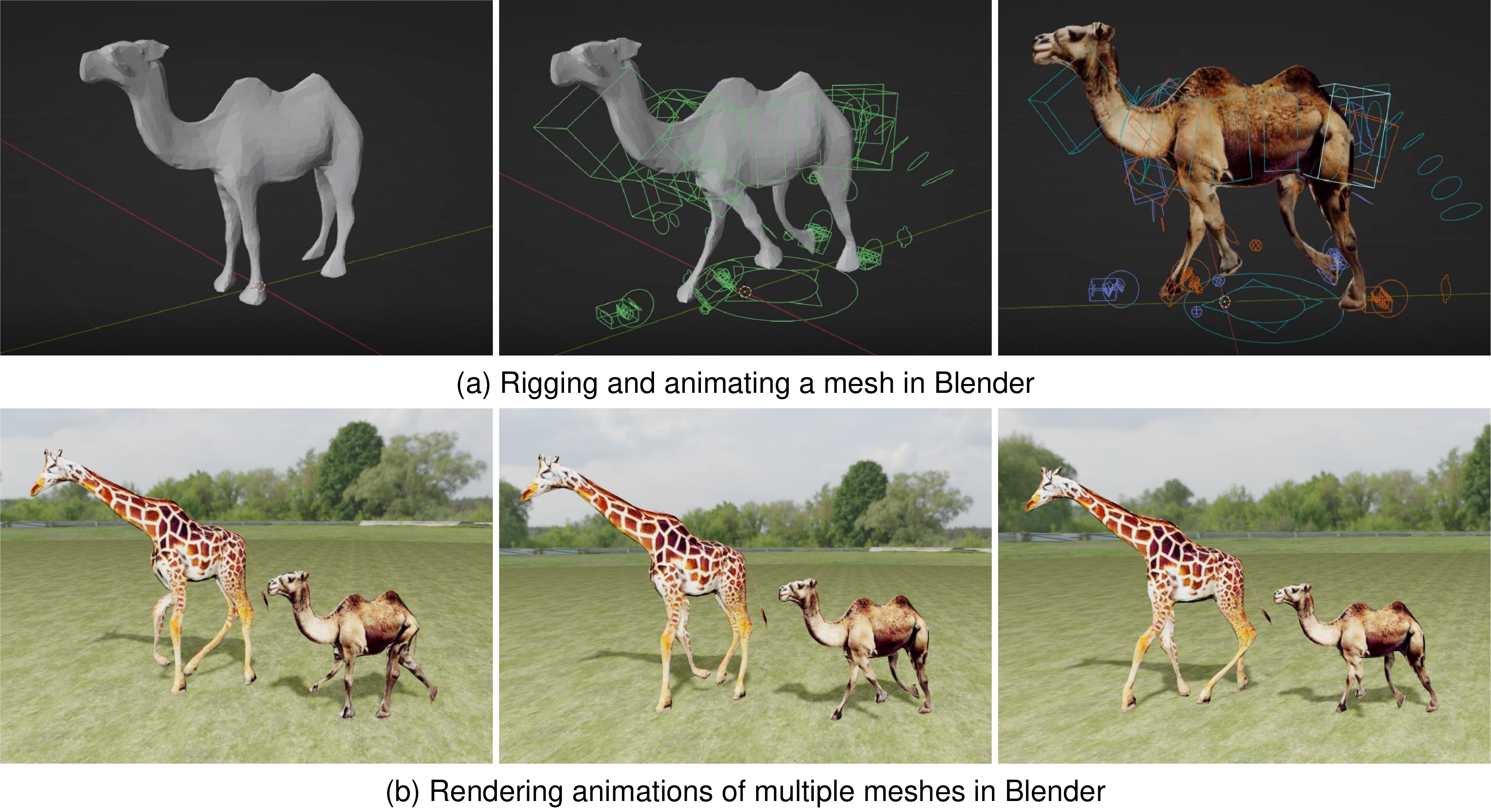}
	\caption{We can directly rig and animate our generated mesh in Blender (a). We can render an animation by using the textured meshes generated by our DreamMesh in Blender, such as a giraffe and a camel walking on a grassy field (b).}
	\label{fig:mesh-animate}
\end{figure}

Unlike previous methods that predominantly revolve around implicit 3D scene representation of density-based geometry with undefined surface boundaries, our work frames text-to-3D generation based on a completely explicit 3D scene representation of triangle meshes. This enables us to generate high-quality meshes with a clean and well-organized topology featuring neatly arranged vertices, edges, and faces. Consequently, these generated meshes can be seamlessly integrated into the traditional graphics rendering pipelines, enabling effortless deployment in standard visualization and animation workflows. Herein, we showcase two types of applications of our DreamMesh in the 3D rendering pipeline. First, the textured mesh generated by DreamMesh can be easily rigged and animated in Blender (see Figure \ref{fig:mesh-animate} (a)). Second, we can use DreamMesh to generate multiple meshes and put them into a 3D scene to render an animation by Blender (see Figure \ref{fig:mesh-animate} (b)). The rigging and animating application videos are also showcased on our project page (\href{https://dreammesh.github.io}{https://dreammesh.github.io}).

\section{More Qualitative Comparisons with Baselines}
\begin{figure}[t]
	\centering
	\includegraphics[width=1.0\textwidth]{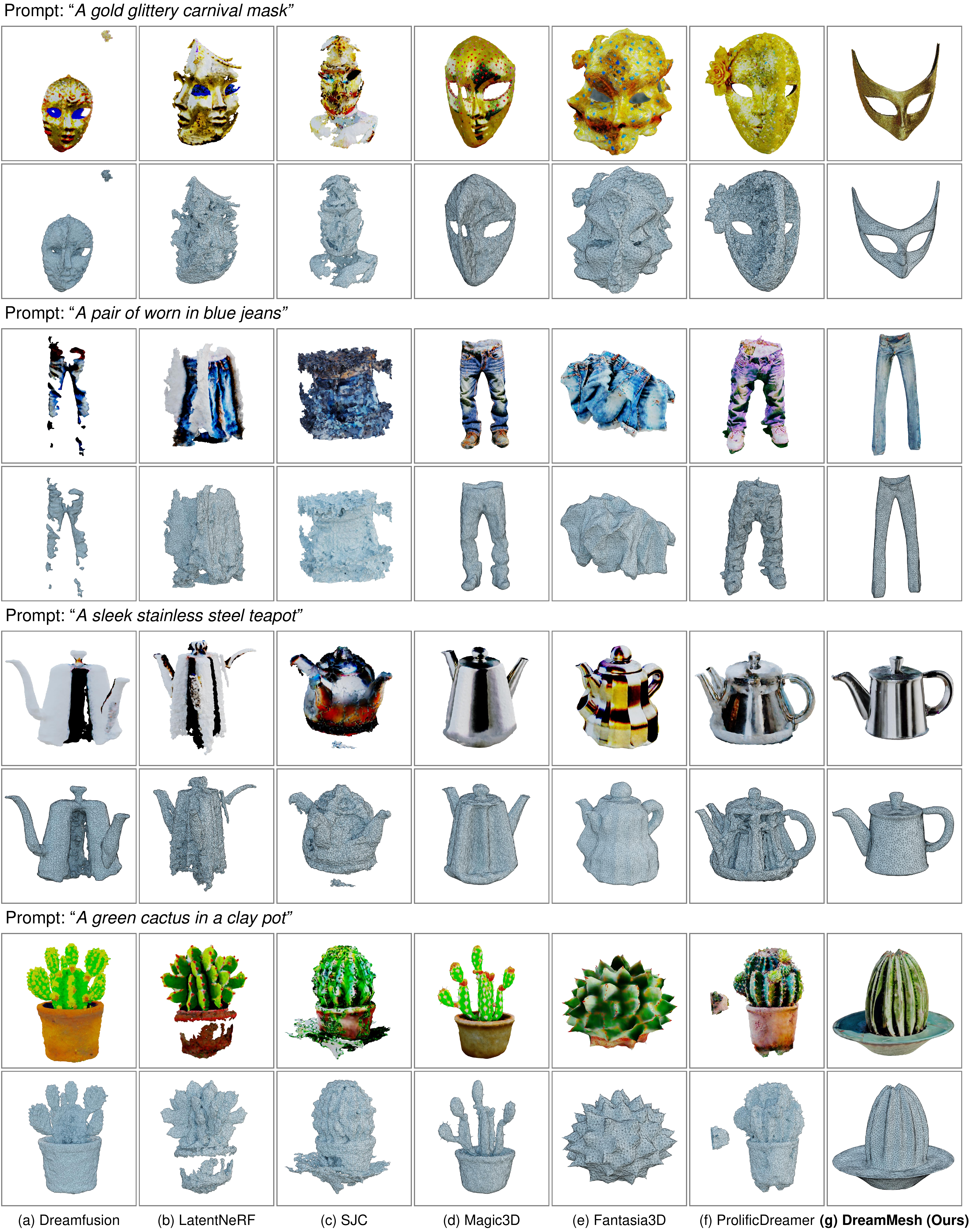}
	\caption{Additional qualitative comparison of texture and wireframe results (rendering in Blender) between our DreamMesh and other baseline methods.}
	\label{fig:add1}
\end{figure}

Here, we provide more qualitative comparisons with baselines on the T$^3$Bench benchmark \cite{he2023t3bench}. T$^3$Bench has released the textured mesh generated by all baseline methods ~\cite{poole2022dreamfusion, metzer2022latent-nerf, wang2022SJC, lin2022magic3d, fantasia3d, wang2023prolificdreamer} on their GitHub repository. We compare them with the generated textured mesh by our DreamMesh. For a fair comparison, we use Blender to render images and wireframes from these meshes under the same lighting environment. Figure \ref{fig:add1} showcases some comparison results. It is easy to see that baseline methods tend to generate distorted geometry and unrealistic textures. In contrast, our DreamMesh can generate higher-quality textured meshes characterized by clean mesh topology and realistic textures.

\end{document}